\documentclass[sigconf]{aamas}  % do not change this line!

\usepackage{subcaption}
\usepackage{tikz}
\usetikzlibrary{calc,positioning,shapes,shadows,arrows,fit}
\usepackage{array}
\usepackage{booktabs}
\usepackage{enumitem}
\setlist[itemize]{leftmargin=3mm}

\setcopyright{ifaamas}  % do not change this line!
\acmDOI{doi}  % do not change this line!
\acmISBN{}  % do not change this line!
\acmConference[AAMAS'18]{Proc.\@ of the 17th International Conference on Autonomous Agents and Multiagent Systems (AAMAS 2018), M.~Dastani, G.~Sukthankar, E.~Andre, S.~Koenig (eds.)}{July 2018}{Stockholm, Sweden}  % do not change this line!
\acmYear{2018}  % do not change this line!
\copyrightyear{2018}  % do not change this line!
\acmPrice{}  % do not change this line!

\begin{document}

\title{Ordered Preference Elicitation Strategies for Supporting Multi-Objective Decision Making}

% AAMAS: as appropriate, uncomment one subtitle line; check the CFP
%\subtitle{Extended Abstract}
%\subtitle{Industrial Applications Track}
%\subtitle{Socially Interactive Agents Track}
%\subtitle{Blue Sky Ideas Track}
%\subtitle{Robotics Track}
%\subtitle{JAAMAS Track}
%\subtitle{Doctoral Mentoring Program}

%\subtitlenote{The full version of the author's guide is available as \texttt{acmart.pdf} document}

%% example of author block for camera ready version of accepted papers: don't use for anonymous submissions
%
\author{Luisa M Zintgraf}
\affiliation{\institution{Vrije Universiteit Brussel \\ University of Oxford}}
\author{Diederik M Roijers}
\affiliation{\institution{Vrije Universiteit Brussel \\ Vrije Universiteit Amsterdam}}
\author{Sjoerd Linders}
\affiliation{\institution{City of Amsterdam, \\ Signal Control Design and Analysis}}
\author{Catholijn M Jonker}
\affiliation{\institution{Delft University of Technology}}
\author{Ann Now\'e}
\affiliation{\institution{Vrije Universiteit Brussel}}

\begin{abstract}  
In multi-objective decision planning and learning, much attention is paid to producing optimal solution sets that contain an optimal policy for every possible user preference profile.
We argue that the step that follows, i.e, determining which policy to execute by maximising the user's intrinsic utility function over this (possibly infinite) set, is under-studied.
This paper aims to fill this gap.
We build on previous work on Gaussian processes and pairwise comparisons for preference modelling, extend it to the multi-objective decision support scenario, and propose new ordered preference elicitation strategies based on ranking and clustering.
Our main contribution is an in-depth evaluation of these strategies using computer and human-based experiments.
We show that our proposed elicitation strategies outperform the currently used pairwise methods, and found that users prefer ranking most. Our experiments further show that utilising monotonicity information in GPs by using a linear prior mean at the start and virtual comparisons to the nadir and ideal points, increases performance.
We demonstrate our decision support framework in a real-world study on traffic regulation, conducted with the city of Amsterdam.
\end{abstract}

% AAMAS: the ACM CCS are not needed within AAMAS papers
%%
%% The code below should be generated by the tool at
%% http://dl.acm.org/ccs.cfm
%\begin{CCSXML}
%<ccs2012>
%  <concept_id>10010520.10010553.10010562</concept_id>
%  <concept_desc>Computer systems organization~Embedded systems</concept_desc>
%  <concept_significance>500</concept_significance>
%  <concept_id>10010520.10010575.10010755</concept_id>
%  <concept_desc>Computer systems organization~Redundancy</concept_desc>
%  <concept_significance>300</concept_significance>
%  <concept_id>10010520.10010553.10010554</concept_id>
%  <concept_desc>Computer systems organization~Robotics</concept_desc>
%  <concept_significance>100</concept_significance>
%  <concept_id>10003033.10003083.10003095</concept_id>
%  <concept_desc>Networks~Network reliability</concept_desc>
%  <concept_significance>100</concept_significance>
%</ccs2012>  
%\end{CCSXML}
%
%\ccsdesc[500]{Computer systems organization~Embedded systems}


\maketitle

\section{INTRODUCTION}

\begin{figure}
\includegraphics[width=\columnwidth]{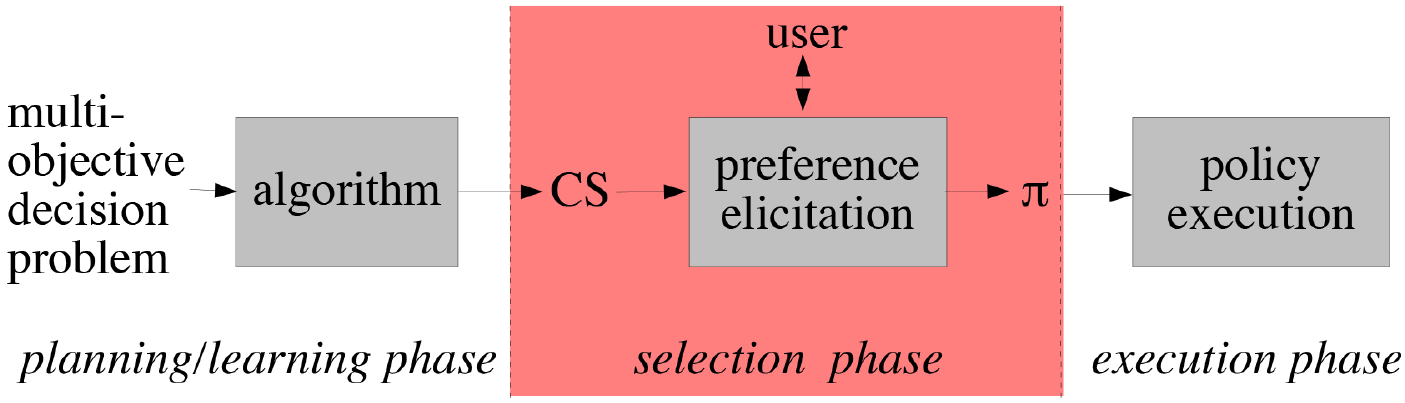}
\centering
\caption{\small \textbf{Decision support scenario.} In the planning or learning phase a coverage set (CS) for a multi-objective decision problem is produced. In the selection phase, the user selects the policy that maximises his or her utility by interacting with an algorithm for preference elicitation. Finally, the selected policy is executed. In this paper, we focus on the selection phase.}
\vspace{-0.3cm}
\label{fig:decision-support-scenario}
\end{figure}

\emph{Understanding what humans want} is an integral part of artificial intelligence (AI), and of central importance when using AI to assists humans in making decisions. Consider tasks like picking which film to watch, or deciding on a route for a road-trip through Europe: the amount of options to choose from is often too large for a human to iterate through, making the search for the best option a possibly cumbersome process. In this case, AI can support -- and accelerate -- the user's decision-making. To this end, the AI system needs to learn about the user's preferences efficiently by \emph{asking the right questions} and guide the search by \emph{generalising to new options}.

AI-supported human decision making is used, e.g., in recommender systems \cite{adomavicius2015context, rashid2002getting, christakopoulou2016towards} or negotiating agents \cite{jonker2007agent, haynes1997automated}. In this paper, we focus on decision support in the context of multi-objective decision making \cite{greco2005multiple, roijers2017modem}, where each option has several attributes that influence the user's preferences.
The AI subfield of multi-objective decision theoretic planning and learning \cite{barrett2008learning,lizotte2012linear,mossalam2016multi,parisi2016multi,van2014multi,wang2013hypervolume,white1980solution,wiering2007computing} studies complex multi-objective decision problems. Such studies typically focus on producing so-called coverage sets, i.e., a set that contains an \emph{optimal policy for every possible user preference profile} with respect to the different objectives.
It is commonly (explicitly or implicitly) assumed that user preferences can be modelled by a utility function that is monotonically increasing in all objectives, leading to a possibly infinitely sized Pareto front as coverage set. However, the selection problem that follows (fig \ref{fig:decision-support-scenario}), i.e., \emph{selecting the policy the user likes best}, is typically left open. Since we do not have direct access to the user's \emph{intrinsic utility function}, this step is far from trivial. We argue, in accordance with the utility-based approach to multi-objective decision making \cite{roijers2013survey, zintgraf2015quality}, that the selection phase is an integral part of solving a multi-objective decision problem, and that doing this suboptimally can be detrimental to user utility. Therefore, we believe that algorithms for the selection phase should be seen as an essential part of multi-objective decision-theoretic planning and learning. These should learn about a specific user's utility with respect to the objectives, and find a single (approximately) optimal policy in terms of user utility which can be executed. In this paper we propose and analyse methods to do so, under two main considerations: how to elicit the user's preferences, and how to model these in the multi-objective setting.
To find the policy that the user likes best, we take an \emph{active learning} approach, where we alternate between updating the model of the user's utility and querying the user for feedback by using \emph{relative} feedback queries. Relative comparisons are a natural way for humans to express preferences (as discussed in sec \ref{sec:background-preferences}). To model the user's preferences, we utilise Gaussian processes (GPs), since Bayesian methods like these work well with little data (and interactions with the user are a scarce resource, see sec \ref{sec:background-gp}). We build on a method that uses a novel likelihood for pairwise comparisons in GPs \cite{chu2005preference} (sec \ref{sec:background-gp}), and were further inspired by the active learning setting of \citet{brochu2008active}, who iterate between asking the user for feedback via a pairwise comparison, updating the GP with this information, and selecting the next query (sec \ref{sec:background-active}).

In this paper we extend their approach to other query types, like asking the user to \emph{rank} or \emph{cluster} items (sec \ref{sec:contribution-queries}). Our main contribution is an in-depth evaluation of these in synthetic experiments (sec \ref{sec:exp1-quality}), and in a user study (sec \ref{sec:exp2-user}). Our results show that ranking queries lead to better utility models, and are preferred by humans over the pairwise and clustering approach.
We further propose to utilise the monotonicity assumption of multi-objective decision problems by using a linear prior mean for the GP, and virtual comparisons (sec \ref{sec:contribution-multi}). We show experimentally that this indeed improves performance (sec \ref{sec:exp1-monotonicity}). We also found that while using a linear prior mean leads to a large performance increase at the beginning, it is essential to turn off the prior after some initial data has been collected, to not restrict the GP too much.
Finally, we demonstrate the implementation of our method in a real-world example; a project in collaboration with the municipality for traffic regulations of Amsterdam (sec \ref{sec:exp3-traffic}).

The source code for our methods and experiments can be found at \url{https://github.com/lmzintgraf/gp_pref_elicit}.

\section{BACKGROUND} \label{sec:background}

In this section we introduce the framework of multi-objective decision making and how user utilities are formalised within it. We then introduce Gaussian processes, and how to use them with pairwise comparative judgements and active learning.

\subsection{Multi-Objective Decision Making} \label{sec:background-modem}

Much multi-objective decision making takes place in the context of the \emph{decision support scenario} \cite{roijers2013survey} (fig \ref{fig:decision-support-scenario}), which applies to multi-objective decision-theoretic planning and learning \cite{roijers2013survey}, but also to multi-objective (heuristic) optimisation with, e.g., evolutionary algorithms \cite{coello2007evolutionary}. 
The ultimate goal in this scenario is to \emph{maximise user utility} by executing the policy a specific user likes best. However, users typically cannot express their preferences directly, and thus \emph{a priori} scalarisation of the objectives is not possible. In this case, we need an approach that delivers a \textit{coverage set} (CS) of \emph{possibly optimal} solutions. We refer to this as the planning/learning phase. The selection phase that follows, i.e, finding the policy from this set that maximises a specific user's utility, is the focus of this paper.

We call a single solution in a multi-objective decision problem a \emph{policy} $\pi \in \Pi$ and denote its value as $\mathbf{v}\in\mathbb{R}^d$, where $d\ge2$ is the number of objectives and $\Pi$ the set of all possible policies. In the context of \emph{multi-objective Markov decision problems} (MOMDPs) \cite{white1980solution} for example, a policy's value is its expected cumulative reward in each objective.
Following \citet{roijers2013survey}, we assume that the user has as an \emph{intrinsic utility function} 
\begin{equation} \label{eq:utility-function}
u:\mathbf{v}\mapsto u_\mathbf{v} 
\end{equation}
that maps a multi-dimensional policy value to a scalar $u_\mathbf{v} \in \mathbb{R}$ according to the user's preferences (a common assumption in general literature on decision making and preference elicitation \cite{clemen2013making, furnkranz2010preference}). The higher this value, the higher the user's preference for the policy. Since all objectives are desirable, $u$ is monotonically increasing in all objectives.
Given this monotonicity property, the solution set is a \textit{Pareto coverage set (PCS)}, i.e., Pareto front, that contains for any allowed policy $\pi'\in\Pi$ with value $\mathbf{v}'$ a policy that has a greater or equal value in all objectives \cite{roijers2013survey}. Consequently it contains an optimal solution for any user preference profile,
\begin{equation}
\forall u \ \exists (\mathbf{v}\!\!\in\!\text{PCS}) ~~:~~ \forall \mathbf{v}' \in \Pi: u(\mathbf{v}) \le u(\mathbf{v}') \ .
\end{equation}
When $\Pi$ is an infinite set of policies (e.g., the set of all possible stochastic policies for an MOMDP), the PCS can contain infinitely many policies. However, due to a result by \citet{vamplew2009constructing}, it is typically possible to construct a PCS by mixing policies from a much smaller solution set, alleviating the necessity to compute the PCS explicitly.
More precisely, we can use the \emph{convex coverage set (CCS)} which contains non-dominated policies whose values satisfy
\begin{equation}
\forall \mathbf{w}\in\mathbb{R}^d \exists \mathbf{v} \in \text{CCS}: \forall \mathbf{v}' \in \Pi: \mathbf{w}^\top\mathbf{v}' \le \mathbf{w}^\top\mathbf{v}  \ ,
\end{equation}
with non-negative weights $\mathbf{w}$ that sum to $1$.
This insight is important, because the CCS is typically much smaller than the PCS. E.g., in MOMDPs with a finite number of states and actions there is a CCS that is a finite set of deterministic stationary policies. The PCS for MOMDPs is then a piecewise linear surface comprised of (infinitely many) mixture policies that have policies from the CCS as constituent policies. A mixture policy is a stochastic mixture of two or more policies, and the value of this policy is the linear combination of the adjacent policies (see \citet{vamplew2009constructing}).
In the decision support scenario, we want to maximise the utility function $u$, \textit{restricted to the mixtures of CCS policy values} as input. To this end, we need to elicit preference information from the user.

\subsection{Preference Elicitation} \label{sec:background-preferences}

We distinguish two ways of querying user preferences: scoring of items (absolute feedback), or comparisons between items (relative feedback). Expressing preferences in absolute terms is more difficult for humans and prone to errors, as numbers are an unnatural way to express preferences \cite{tesauro1988connectionist} (consider ``I like this film $0.4$ much'') and the values might change over time \cite{siegel1956nonparametric} with the users mood \cite{forgas1995mood,sirakaya2004role} which may depend on something seemingly trivial like the weather. Relative feedback \cite{tesauro1988connectionist, zoghi2014relative} is easier for humans to express (consider ``I prefer film A over B''), and is typically more consistent over time \cite{kingsley2006preference}. We therefore focus on relative feedback.

When a user compares policies $\mathbf{v}_1$ and $\mathbf{v}_2$, we assume there are true utility values $u(\mathbf{v}_1)$ and $u(\mathbf{v}_2)$ which the user cannot directly access, but uses indirectly to compare items.
Since personal evaluations by users are likely not always $100\%$ accurate and vary depending on different factors, we model the outcome of a comparison based on \emph{noisy} utility values $u(\mathbf{v})+\varepsilon$, where $\varepsilon\sim\mathcal{N}(0, \sigma^2)$ is Gaussian noise with zero mean and unknown variance (following \cite{chu2005preference, brochu2008active}).
The observations we get from the user thus stem from the comparison of the true utility values, contaminated with noise,
\begin{equation} \label{eq:noisy-comparison}
u(\mathbf{v}_1) + \varepsilon_{\mathbf{v}_1} > u(\mathbf{v}_2) + \varepsilon_{\mathbf{v}_2},
\end{equation}
denoted as $\mathbf{v}_1\succ \mathbf{v}_2$ (meaning the user prefers $\mathbf{v}_1$). We want to use such relative preferences to find the policy $\pi^*$ whose value $\mathbf{v}^*$ maximises the user's utility function $u$. In the next section we introduce an approximation method that is suitable for this task.

\subsection{Gaussian Processes for Pairwise Feedback} \label{sec:background-gp}

A key objective when asking for feedback is to not bother the user with too many queries -- otherwise, the user might quit using our system before its goals are achieved. We therefore need methods that work well with little data and take the uncertainty about the user's preferences into account. Bayesian optimisation techniques, like Gaussian processes, are particularly suited for tasks where there is little available data \cite{deisenroth2015gaussian}.

Gaussian processes (see \cite{rasmussen2006gaussian} for a general introduction) are used to approximate functions, and can be seen as an infinite-dimensional extension of a multivariate Gaussian distribution. A GP approximates a function and captures its uncertainties by assigning to each point in its domain a normal distribution -- the mean value reflecting the expected value, and the variance reflecting the uncertainty about the function value at that point. The Gaussian process is fully specified by a mean function $m$ and a kernel $k$,
\begin{equation}
u(\mathbf{v}) \sim GP(m(\mathbf{v}), k(\mathbf{v},\mathbf{v}')) \ .
\end{equation}
Before querying the user for information, the GP is initialised by defining a prior mean function and the kernel function. Common choices are the zero function $m(x)=0$ for the prior mean and the squared exponential kernel which we also use here; an alternative prior mean function is discussed in section \ref{sec:multi-monotonicity}.

The prior belief $P(u)$ (defined by the GP) about the utility function $u$ is updated in light of new data and its likelihood $P(D|u)$ using Bayes' rule
$ P(u|D) \propto P(u) P(D|u)$ .
The posterior $P(u|D)$ is the updated belief about the user's utility, and the current approximation of the utility function.
The data $\mathcal{D}$ in our case is given in terms of pairwise comparisons (see equation (\ref{eq:noisy-comparison})),
\begin{equation} \label{eq:dataset}
\mathcal{D} = \{ \mathbf{v}_m \succ \mathbf{v}'_m \}_{m=1}^M.
\end{equation}
Chu and Ghahramani  \cite{chu2005preference} introduce a probit likelihood for such noisy pairwise comparisons. 
Given this likelihood and a Gaussian process prior, the posterior is analytically not tractable and has to be approximated. Following \cite{brochu2010tutorial}, we use Laplace approximation.

To approximate the user's utility with a GP, we take an active learning approach, meaning we alternate between updating the GP (approximating the posterior) and querying the user for more information (based on our current approximation). In the next section, we describe how to select the next question for the user.

\subsection{Active Learning} \label{sec:background-active}

Chu and Ghahramani \cite{chu2005preference} show that GPs with the above defined pairwise likelihood work well on fixed datasets, where learning is done off-line and (during learning) we do not have influence on the information that is collected from the user. Here instead, we are interested in an active learning setting, where the queries for new information are based on the current information and belief about the user's preferences. Functions that select a next item given a Gaussian process are called acquisition functions. These typically try to balance exploration (querying the user about items with high uncertainty in user utility) and exploitation (suggesting items to the user that have high expected utility).
Following \cite{brochu2008active,brochu2010tutorial}, we use the \textit{expected improvement} acquisition function \cite{lizotte2008practical} to, at each step, select an item about which the user is queried about next.

\section{MULTI-OBJECTIVE DECISION SUPPORT WITH GAUSSIAN PROCESSES} \label{sec:contribution-multi}

To use GPs for optimising user utility in multi-objective decision support, we restrict the domain of the acquisition function (\ref{sec:contribution-multi-domain}) and utilise monotonicity information about the user's preferences (\ref{sec:multi-monotonicity}).

\subsection{Input Domain} \label{sec:contribution-multi-domain}

The input domain of the user's utility function, as well as the GP we use to approximate it, is the $d$-dimensional hypercube defined by the (possibly unknown) minimum and maximum possible values of the $d$ objectives. In the multi-objective planning or learning phase (fig \ref{fig:decision-support-scenario}), a \emph{coverage set} is produced which contains an optimal solution for every possible user. This will typically be an infinite subset of the $d$-dimensional hypercube, since trade-offs between the objectives constrain the solution set. On this set, the user's utility function can have local maxima.
Since we aim to maximise the user's utility on this solution set, we restrict the acquisition function to this set. This way, we only present achievable solutions to the user. Doing otherwise could bias the user, give rise to unrealistic expectations, and we might waste time asking the user about items that are not feasible.

\subsection{Utilising Monotonicity Information} \label{sec:multi-monotonicity}

In the multi-objective decision support scenario, we assume the user's utility function to be monotonically increasing in each objective. This information can be used in the GP approximation.

\subsubsection{Prior Mean}
If no information about the function that is approximated is known a priori, the most commonly used GP prior mean is the zero function (sec \ref{sec:background-gp}). However, because we have a monotonic utility function, we want to use a better heuristic as a prior. We propose a \emph{linear} prior mean function with equal weights for all objectives for this heuristic. Equal weights implies that the user would care equally about all objectives, and linearity means that the user's utility increases linearly in the value of each objective. Note that a linear prior mean function does not imply that the GP can only model linear functions; the shape of the functions is determined by the kernel.

In Section \ref{sec:exp1-monotonicity} we show that adding such a heuristic prior is highly useful when not many queries have been posed to the user, but can hinder the GP when more data becomes available. Hence, we propose to remove this heuristic after an initial number of queries.

\subsubsection{Virtual Comparisons}
We optimise the user's utility on the set of achievable and possibly optimal solutions, i.e., the PCS. Around the PCS however, we can add virtual comparisons (which are not shown to the user) that include items that lie \emph{outside} the PCS, to enforce monotonicity.
We propose to compare each item that is selected by the acquisition function to the \emph{minimal} and \emph{maximal item} (also called the nadir and ideal points). These are the vectors that contain the minimal (resp. maximal) value for each objective, independent of the other objective values. We hypothesise that because this extra information enforces some monotonicity for the GP, it will lead to better approximations.

\section{PREFERENCE ELICITATION STRATEGIES} \label{sec:contribution-queries}

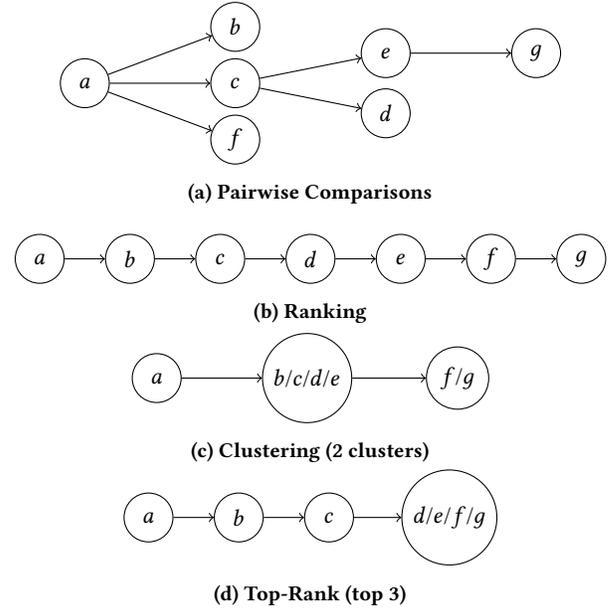
\begin{figure}[t]
%
% PAIRWISE
\begin{subfigure}[b]{\linewidth}
\centering
  \begin{tikzpicture}[every node/.style={minimum size=0.65cm}]
  \node(0) at (0,0)		[shape=circle,draw] 	{$a$};
  \node(1) at (2,0.75)	[shape=circle,draw]		{$b$};
  \node(5) at (2,0)		[shape=circle,draw]		{$c$};
  \node(2) at (2,-0.75)	[shape=circle,draw]		{$f$};
  \node(3) at (4,0.4)	[shape=circle,draw]		{$e$};
  \node(4) at (4,-0.4)	[shape=circle,draw]		{$d$};
  \node(6) at (6,0.4)	[shape=circle,draw]		{$g$};
  \path [->]
    (0)      edge	node [above]  {}     (1)
    (0)      edge	node [above]  {}     (2)
    (5)      edge	node [above]  {}     (3)
    (0)      edge	node [above]  {}     (5)
    (5)      edge	node [above]  {}     (4)
    (3)      edge	node [above]  {}     (6);
  \end{tikzpicture}
  \caption{Pairwise Comparisons}
  \label{fig:order-pairwise}
\end{subfigure}
%
% RANKING
\begin{subfigure}[b]{\linewidth}
\centering
  \begin{tikzpicture}[every node/.style={minimum size=0.65cm}]
  \node(0) at (0,0)		[shape=circle,draw] 	{$a$};
  \node(1) at (1.2,0)	[shape=circle,draw]  	{$b$};
  \node(2) at (2.4,0)	[shape=circle,draw]  	{$c$};
  \node(3) at (3.6,0)	[shape=circle,draw] 	{$d$};
  \node(4) at (4.8,0)	[shape=circle,draw] 	{$e$};
  \node(5) at (6,0)		[shape=circle,draw] 	{$f$};
  \node(6) at (7.2,0)	[shape=circle,draw] 	{$g$};
  \path [->]
    (0)		edge	node [above]  {}     (1)
    (1)		edge	node [above]  {}     (2)
    (2)		edge	node [above]  {}     (3)
    (3)		edge	node [above]  {}     (4)
    (4)		edge	node [above]  {}     (5)
    (5)		edge	node [above]  {}     (6);
  \end{tikzpicture}
  \caption{Ranking}
  \label{fig:order-ranking}
\end{subfigure}
%
% CLUSTERING
\begin{subfigure}[b]{\linewidth}
\centering
  \begin{tikzpicture}[every node/.style={minimum size=0.65cm}]
  \node(0) at (0,0)[shape=circle,draw] {$a$};
  \node(1) at (2,0)[shape=circle,draw] {$b$/$c$/$d$/$e$};
  \node(2) at (4,0)[shape=circle,draw] {$f$/$g$};
  \path [->]
    (0)      edge                 node [above]  {}     (1)
    (1)      edge                 node [above]  {}     (2);
  \end{tikzpicture}
  \caption{Clustering (2 clusters)}
  \label{fig:order-clustering}
\end{subfigure}
%
% TOP-RANK
\begin{subfigure}[b]{\linewidth}
\centering
  \begin{tikzpicture}[every node/.style={minimum size=0.65cm}]
  \node(0) at (0,0)		[shape=circle,draw] 	{$a$};
  \node(1) at (1.2,0)	[shape=circle,draw]  	{$b$};
  \node(2) at (2.4,0)	[shape=circle,draw]  	{$c$};
  \node(3) at (4,0)[shape=circle,draw] {$d$/$e$/$f$/$g$};
  \path [->]
    (0)      edge                 node [above]  {}     (1)
    (1)      edge                 node [above]  {}     (2)
    (2)      edge                 node [above]  {}     (3);
  \end{tikzpicture}
  \caption{Top-Rank (top 3)}
  \label{fig:order-top-rank}
\end{subfigure}
\caption{\small Possible outcomes of different query types for items $a$-$g$, with utilities $u(a)>...>u(g)$. The arrows represent the preference information expressed by the user (preferred $\rightarrow$ unfavoured). Different elicitation strategies lead to different orderings: full ranking returns a total ordering (b); the other query types typically lead to partial orderings.}
\label{fig:query-types}
\end{figure}

This section introduces the query types we study in this paper; pairwise comparisons (from previous work), and new ranking/clustering strategies. We describe all query types in the following, and show orderings that may result in figure \ref{fig:query-types}.

\begin{itemize}
	\item \textbf{Pairwise Comparisons.}
      \citet{brochu2010tutorial} use an active learning approach with pairwise comparisons: in each step, the user compares two items; the winner of which will be compared to a new item in the next query. Thus after $N$ queries, the user made $N-1$ comparisons, which are used to (sequentially) update the GP. This typically leads to a partial ordering of items (\ref{fig:order-pairwise}); the extreme cases being a total ordering and a degenerate case where the first item is better than all items from subsequent queries.
    \item  \textbf{Ranking.}
        Here, the user is asked to make a full ranking of all inputs, leading to a total ordering over the items (\ref{fig:order-ranking}). The user initially starts with ranking two items, and in each round one more item is added, which has to be sorted into the ranking. I.e., after $N$ queries the user has evaluated $N-1$ items and ordered those as a sorted list. We use the $N-1$ pairs of \emph{successive} items to update the GP.
        Note that this ranking approach leads to the same number of comparisons as pairwise comparisons, but we get a full ordering. In other words, the data quantity is the same (assuming the user does not re-arrange items), but the information quality is higher when doing ranking, compared to pairwise comparisons.
    \item \textbf{Clustering.}
      Instead of requesting a full ranking, we can ask the user to: pick a best item, and put the other items into clusters of decreasing utility (\ref{fig:order-clustering}). Items whose utility values are close are put in the same cluster. The number of clusters can either be pre-determined or defined dynamically by the user. With two clusters, the user's preferences would then be expressed in the form
      \begin{equation}
      x_{best} \succ C_1 \succ C_2 \ ,
      \end{equation}
      where $x_{best}$ is a single best item, and $C_1$ and $C_2$ are disjunct sets whose union contains the remaining items. $C_1 \succ C_2$ means that the items in cluster $C_1$ are better than the ones in $C_2$. We can then use this data in our Gaussian process with the pairwise comparison likelihood by comparing $x_{best}$ to every item in $C_1$, and every item in $C_1$ to every item in $C_2$, giving us $|C_1|+|C_1|*|C_2|$ pairwise comparisons. In general, for any number of clusters, we have \emph{at least} $N-1$ comparisons.
  \item  \textbf{Top-Rank.}
      This query type is a mix of the above: the user is asked to rank the top k items (in figure \ref{fig:order-top-rank} we chose k=3), and all remaining items are put in one cluster. This gives us $N-1$ pairwise comparisons, and is qualitatively between the pairwise and full ranking approach. We test this approach since more information about the \emph{upper region} of the user's utility function might help find the maximum faster.
\end{itemize}

Each experiment starts by asking the user to compare two items, independent of the query types. In each subsequent step, one new item is selected using the acquisition function, and the question asked to the user will depend on the query type. For all ranking/clustering queries (\ref{fig:order-ranking}-\ref{fig:order-top-rank}), we allow the user to re-order the items in each time step, or merge items into the same cluster. The history of previous queries is kept in the dataset that is used to update the GP.
Due to the noise in the utility values and since users are allowed to re-arrange items, it can happen that the relative order of two items is swapped between queries. 
We handle these apparent inconsistencies by just keeping two contradicting pairwise comparisons in the dataset; experiments showed that there is no significant difference to removing these contradictory samples (results not shown).

In the following sections, we evaluate the different query types and study the effect of utilising monotonicity information. We first do this by simulating a user behaving according to a given utility function (sec \ref{sec:exp1-quality}) in order to have access to the true utility values. We then present results from a user study in which we let humans answer queries of different types in a multi-objective decision problem (sec \ref{sec:exp2-user}). Finally, we combine our findings in a real-world application in collaboration with policy makers for traffic regulation in the city of Amsterdam (sec \ref{sec:exp3-traffic}).

\section{EXPERIMENTS I: OPTIMISATION QUALITY} \label{sec:exp1-quality}

In this section, we evaluate our proposed strategies in terms of how well we can optimise the user's utility, i.e., how \emph{close} and how \emph{fast} we get to the maximal achievable utility. For this we need access to the ground truth utility function of the user. We therefore use a virtual user whose utility function we know (read: define). We assess utilising monotonicity information in section \ref{sec:exp1-monotonicity}, and the different query types in section \ref{sec:exp1-queries}. In the following, we first describe our experimental set-up and how we defined the virtual user's utility function and simulate the decision making process.

\subsection{Virtual Utility Function} \label{sec:exp1-setup}

\begin{figure}
\includegraphics[width=\columnwidth]{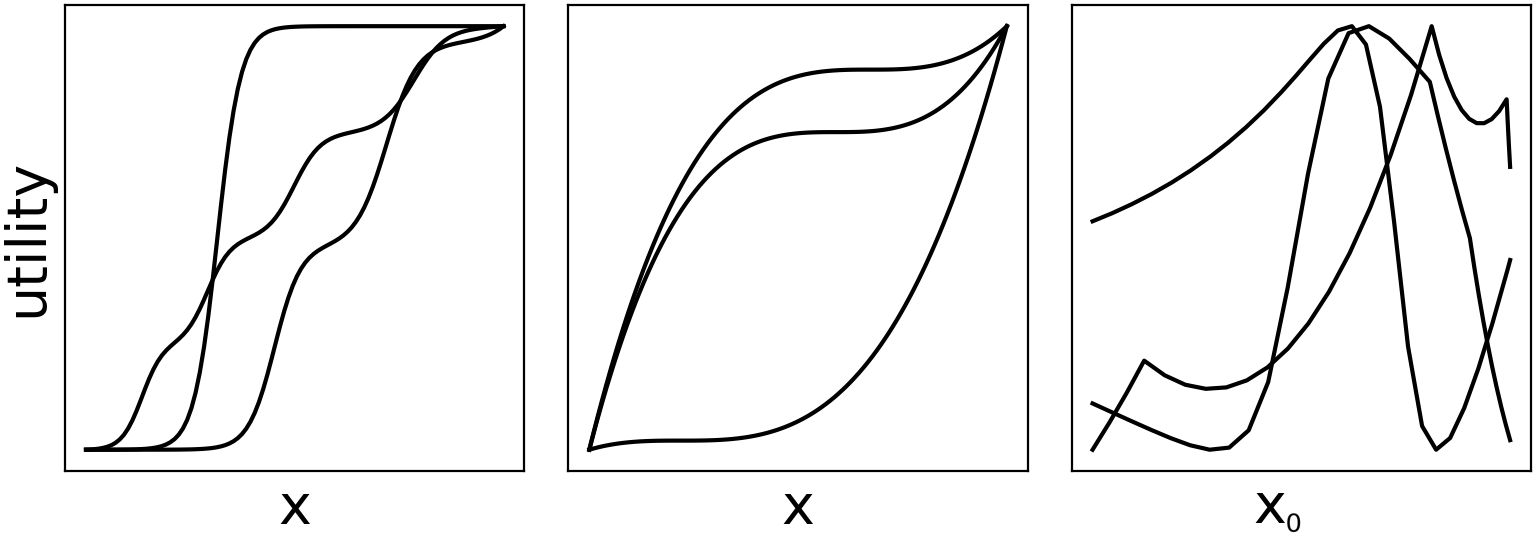}
\caption{\small Examples of stacked sigmoids (\ref{eq:stacked-sigmoid}), polynomials (\ref{eq:polynomial}), and virtual user utility functions (\ref{eq:utility}). On the right, we used $d=2$ and show a slice through the utility function: the mapping of the first objective ($x_0$) to the solution set (PCS).}
\label{fig:general-utility-example}
\end{figure}

First, we define a virtual user by defining a monotonic utility function that represents its preferences, taking the following form
\begin{equation} \label{eq:utility}
	u(\mathbf{v}) = \sum\limits_{i=1}^{d} w_i \cdot f_i(v_i) \ ,
\end{equation}
where $\mathbf{v}$ is the $d$-dimensional value of a policy, $\sum\nolimits_{i=1}^d w_i = 1$, and $f_i(v_i)$ is a monotonically increasing function that maps the $i$-th entry of the vector $\mathbf{v}$ to a scalar value. 
The functions $f_i$ can take two forms, either that of stacked sigmoids
\begin{equation} \label{eq:stacked-sigmoid}
	f_i(x) = \sum\limits_{j-1}^n \frac{1}{1+\exp(-x(a-i)+(b+i))}
\end{equation}
or a polynomial
\begin{equation} \label{eq:polynomial}
	f_i(x) = (c x -1)^3 \ .
\end{equation} 
In our experiments, one run consists of a number of queries asked to the (virtual user). After each query, the GP is updated and a new item is chosen with the acquisition function.
For each run, we first randomly (uniformly) choose the $d$ utility functions $f_i$ we use, as well as the values $a\in[10,50]$, $b\in[1,20]$, $n\in[1,10]$ and $c\in[1,5]$ (separately for each $f_i$). We further randomly generate solution sets (PCSs) which hold all possibly optimal policies, and on which we want to maximise user utility.
The input space is always $x\in[0,1]$, and we normalise the functions $f_i$ individually to map to $[0,1]$ (omitted here in favour of readability) to allow easy comparisons between the results. Figure \ref{fig:general-utility-example} shows examples of stacked sigmoids, polynomials, and general utility functions.

Given this utility function, we simulate users expressing their preferences over items. We start by selecting two starting items $x_1$ and $x_2$ using the acquisition function (note: with zero prior information, these are selected uniformly at random) and then alternate between evaluating the query, updating the GP, and selecting a new item from the acquisition function again. How the queries are evaluated with our virtual utility function is described in the following for the different query types.
\begin{itemize}
	\item \textbf{Pairwise}: At each step, the winner of the previous query $x^*$ and a new item $x_{new}$ are compared. We evaluate the utility function to obtain $u(x^*)$ and $u(x_{new})$, and add noise $\varepsilon^*$ and $\varepsilon_{new}$, drawn i.i.d. from a normal distribution $\mathcal{N}(0, \sigma^2)$. The noisy utility values $u(x^*)+\varepsilon^*$ and $u(x_{new})+\varepsilon^*$ are compared and the result is added to the dataset $\mathcal{D}$. The winner of this comparison is used in the next query (with different noise). 
    \item \textbf{Ranking/Clustering}: At each step, we have $N$ items from previous queries, and a new item $x_{new}$. We evaluate the utility function to obtain $u(x_1),\dotsc,u(x_N),u(x_{new})$, and add noise $\varepsilon_1,\dotsc,\varepsilon_N,$ $\varepsilon_{new}$, drawn i.i.d. from a normal distribution $\mathcal{N}(0, \sigma^2)$. We then produce a ranking or clustering based on the noisy utility values. Note that in each query, we draw new noise values for all items. 
    \item \textbf{Clustering}: To simulate the clustering, we use K-Means (see, e.g., \cite{bishop2006pattern}). There is no general agreement in the literature on how humans perform clustering, especially since this depends heavily on the task itself and prior knowledge. Just like the shape of the utility function itself, using K-Means is therefore a somewhat arbitrary choice based on the intuition of the authors.
\end{itemize}

\subsection{Results: Utilising Monotonicity} \label{sec:exp1-monotonicity}

\begin{figure}
\includegraphics[width=\columnwidth]{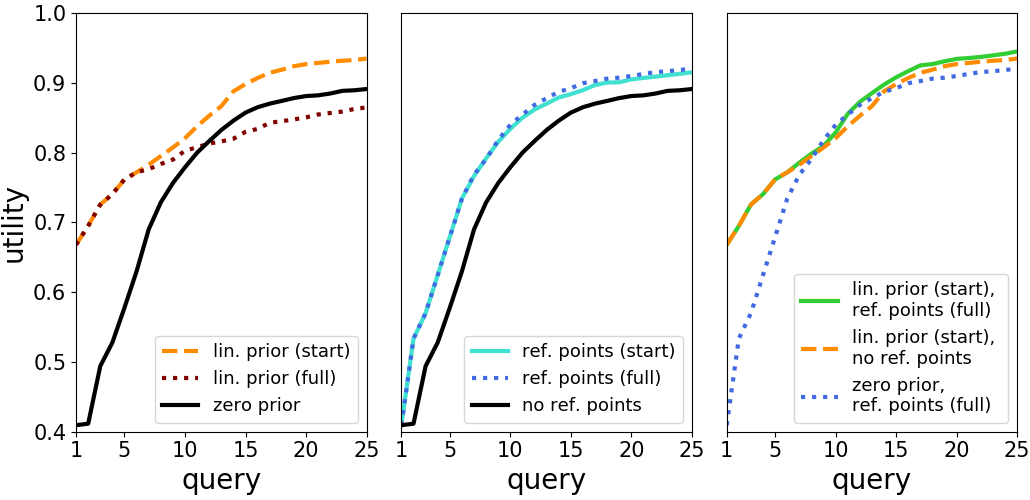}
\centering
\caption{\small \emph{Utilising Monotonicity}: Performance when using different prior functions (left), reference points (middle), and a mix of both (right). We switch from a linear to a zero prior after $5$ queries, and stop adding virtual comparisons after $5$ queries as well. The results are averaged over all query types and 100 runs each. We used $d=5$ objectives and low utility noise with $\sigma=0.01$.} 
\label{fig:exp1-mono}
\end{figure}

Within the multi-decision making framework described in section \ref{sec:background-modem}, we assume that the user's utility function is \emph{monotonic}, i.e., all objectives are desirable. In section \ref{sec:contribution-multi}, we proposed two strategies for utilising this information: adding a linear prior, and adding virtual comparison points. We compare these strategies (and combinations of them) by looking at the reached utility at each query step. Here, we are interested in both how fast good user utility is reached, and the final utility value.

Figure \ref{fig:exp1-mono} shows the results for adding information via the GP prior mean function (left), adding virtual comparisons (middle) and a mix of these (right). Compared to not using any prior information, adding monotonicity information always gives a performance boost. Using a linear prior mean function for the GP at the beginning helps the acquisition function focus on the promising regions, and jump-starts the reached utility. However using a linear prior mean throughout all queries, the performance at some point drops lower than when using a zero prior mean from the start. The same does not hold for reference points. This is because while a linear prior mean function is a heuristic, that might hinder the search process later on, virtual comparisons only enforce monotonicity. The plot on the right shows that a combination of a linear prior at the beginning and virtual comparisons throughout all queries works best.

\begin{figure}
\includegraphics[width=\columnwidth]{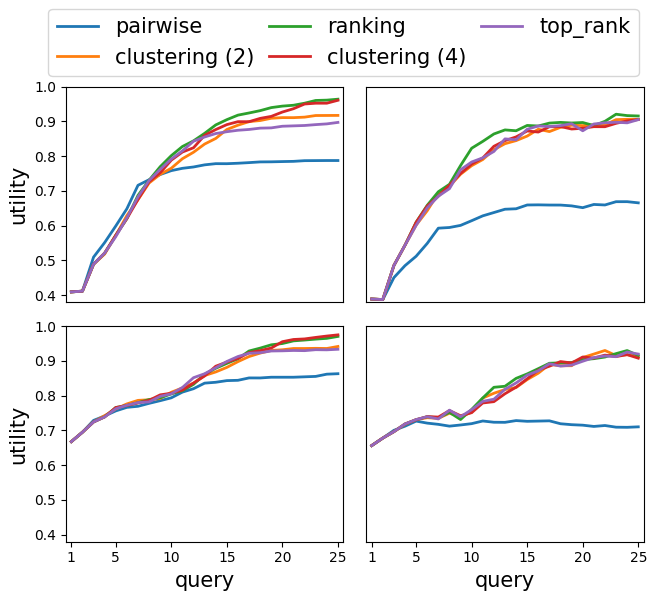}
\centering
\caption{\small \emph{Utility}: This graph shows the reached utility per query. We used $d=5$ objectives and different noise levels $\sigma=0.01$ (left) and $\sigma=0.1$ (right). The results are averaged over 100 runs. The top row shows the performance without using prior information, and the bottom shows the performance when using a linear prior mean function at the beginning, and reference points throughout. Pairwise comparisons are outperformed by other query methods both regarding convergence speed and final reached utility. (Best viewed in colour.)}
\label{fig:exp1-queries}
\end{figure}

\begin{figure}
\includegraphics[width=\columnwidth]{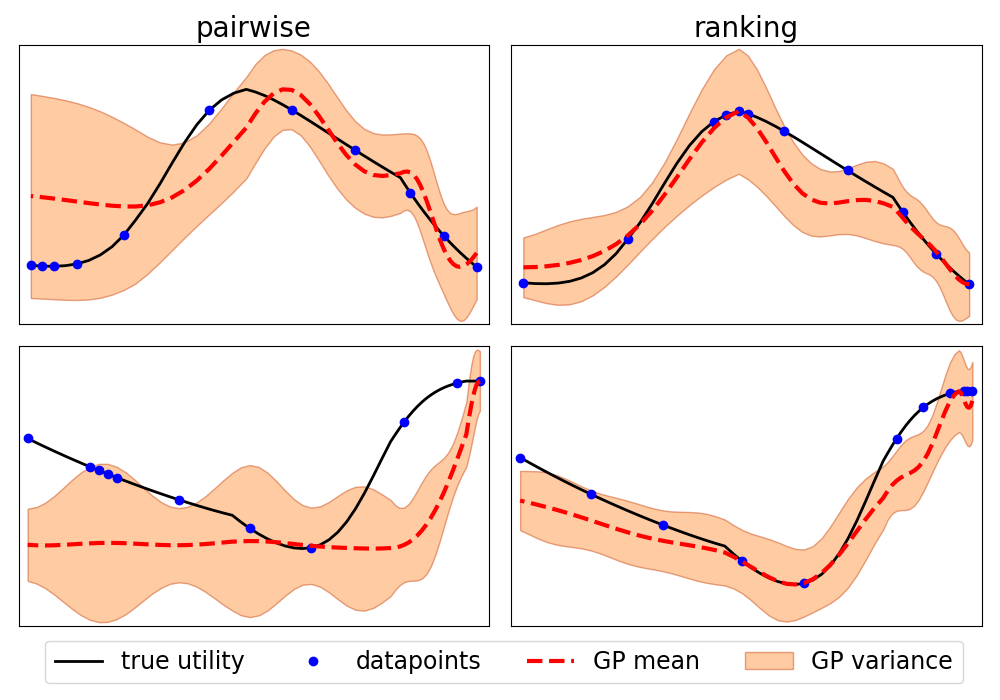}
\centering
\caption{\small \emph{GP Approximation}: This graph shows the approximation of the GP to the true utility function, for $d=2$ objectives and when plotting the first objective against  the utility on the PCS. We show this for two different random seeds (rows) and for the pairwise (left) vs. ranking (right) query, after $5$ queries. As we can see, the ranking query is able to approximate the GP much better. This is due to the different types of orderings (partial vs full) available to the GP on the given datapoints. We hypothesise that a better approximation enables the acquisition function to suggest better points per query, which explains the superior performance of ranking queries.}
\label{fig:exp1-gp-shape}
\end{figure}

\newpage
\subsection{Results: Query Types} \label{sec:exp1-queries}

In this section, we compare our proposed elicitation strategies to the pairwise approach of \cite{brochu2008active} in terms of approximating the utility function, and attaining a high utility. Figure \ref{fig:exp1-queries} shows, per time step, the utility of the currently highest rated item, for $d=5$ objectives and two different noise levels (low/high). The pairwise approach is outperformed by all other methods, especially when there is high noise in the utility function. The full ranking generally performs best, but the difference to clustering and top-rank is not significant when there is a lot of noise in the utility function.

We plot the number of queries against the reached utility, so that each method at each time step has the same datapoints available (since one item is added in each query, regardless of the query type). The difference is in the number of comparisons between those datapoints, which is lowest for the pairwise approach. We note though that the better performance cannot be solely because of the higher number of available comparisons: ranking and clustering methods not only reach higher utility faster, they also converge to a higher utility compared to the pairwise approach.

We believe that the superior performance of ranking and clustering queries stems from having a better approximation to the utility function. While we are only interested in finding the maximum, and not necessarily in closely approximating the user's utility function as a whole, a better approximation helps the acquisition function to make a more informed decision about which item to select next. In figure \ref{fig:exp1-gp-shape} we show the GP approximation to two different utility functions for pairwise queries and ranking after $10$ queries. While both methods are close to the maximum, ranking queries give a better overall approximation. The pairwise method also found an item with near-optimal utility, but since it does not know anything about the relationship between the remaining datapoints, the approximation quality in these areas is poor, and the variance high. One might argue that this is not necessary for finding the maximum element since we are not interested in areas with low utility, but we believe our results give strong indication that a better overall approximation quality can help solve the optimisation problem faster and better.

\section{EXPERIMENTS II: USER STUDY} \label{sec:exp2-user}

In the previous section, we used a virtual user with a utility function we defined ourselves, and we found that ranking queries lead to the best results. In this section, we investigate how real users experience and interact with the system. 

\subsection{Description of User Study}

To assess which query type humans prefer and how well they work in reality, we built a web-interface for multi-objective decision making, including three query types: pairwise, ranking, and clustering (with two clusters). We chose to not test the top-rank method, since this is a mix between ranking and clustering, and we think this might become more relevant for larger studies where many items are compared (and full ranking becomes infeasible).

We chose a decision making scenario over job offers  (as previously used in, e.g., \cite{pommeranz2012designing}). Each job offer has three attributes: the number of days per week on which the employee can work from home, the salary, and the probation time in months. We randomly generated $50$ job offers with these attributes. We then defined a mathematical utility function over job offers, and selected three pairs of jobs with low utility to present in the initial query. We used a zero prior for the GP, and used the acquisition function as described in section \ref{sec:background-active} to select a new job offer at each step.

Each user that participated in our experiment was asked to answer queries about job offers for all query types, according to a \emph{persona description}, in which we described the utility function over jobs in simple words. The participants were given one minute to answer queries of each type. At the end, the user was asked to complete a survey about his or her experience. We collected $50$ responses for this user study via Amazon Mechanical Turk.
When running the experiments, we randomised the order in which the user took the different experiments (i.e., query types) as well as the pair of starting jobs.

\subsection{Results}

\begin{figure}
\includegraphics[width=\columnwidth]{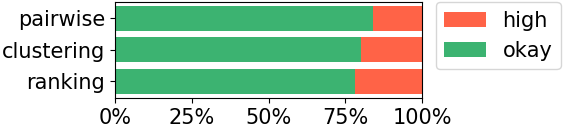}
\centering
\caption{\small \emph{Effort}: For each query type, the participant was able to choose between the options 'effort was okay', and 'too much effort'. Users typically found the effort acceptable for all query types.}
\label{fig:user-study-effort}
\end{figure}

\begin{figure}
\includegraphics[width=\columnwidth]{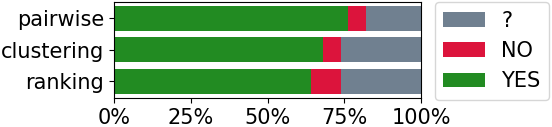}
\centering
\caption{\small \emph{Understanding}: We asked the users whether they felt like the algorithm understood their preferences and suggested better jobs over time. Most people thought it did.}
\label{fig:user-study-understanding}
\end{figure}

\begin{figure}
\includegraphics[width=\columnwidth]{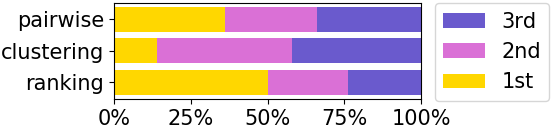}
\centering
\caption{\small \emph{Preference}: We asked the users to rank the query types, according to what they would prefer as a tool to support them in a decision-making situation. Most users preferred ranking.}
\label{fig:user-study-preference}
\end{figure}

\begin{figure}
\includegraphics[width=\columnwidth]{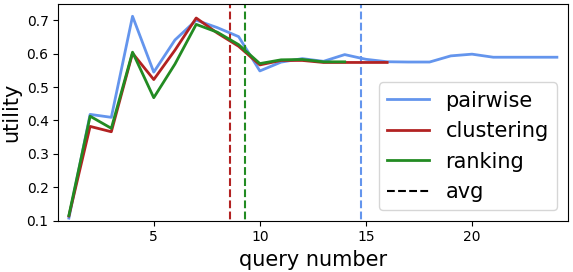}
\centering
\caption{\small \emph{Utility}: This graph shows the reached utility over time, according to the utility function we defined and described to the participants in words. All query types have similar performance. We can also see that the average number of answered queries in the pairwise setting is about 15, and 8-10 in the clustering/ranking scenario within the given minute.}
\label{fig:user-study-utility}
\end{figure}

After participating in our experiment, we asked users about the perceived effort per query, whether they think the algorithm understood their preferences, and which query type they prefer for a decision-making tool. The results are shown in figures \ref{fig:user-study-effort} - \ref{fig:user-study-preference}. We also logged the user's responses to the queries and show the average reached utility per step (according to the true utility function we described to the user in words) in figure \ref{fig:user-study-utility}.

The results show that the majority of participants found the effort to be acceptable across all queries (fig \ref{fig:user-study-effort}), and ranking queries were rated as high effort slightly more often. Most users felt that the algorithm understood their preferences and suggested better jobs over time (fig \ref{fig:user-study-understanding}), and they felt most often so for the pairwise query. Despite these results, pairwise comparisons was not the query type that users prefer for a decision-making tool (fig \ref{fig:user-study-preference}). To our surprise, the ranking query was rated $1$st most often, and $3$rd least often. It further surprised us that clustering was consistently ranked lowest, and that users did not seem to like it. In terms of reached utility (fig \ref{fig:user-study-utility}), all query types show similar performance. The figure also shows the average number of queries answered in one minute per experiment type, and we see that the users were able to answer on average $15$ pairwise queries, and $8$-$10$ for clustering/ranking. 

In conclusion, users prefer the ranking query type in this one-minute experiment, although it requires slightly more effort and the users felt less often understood by this method than the other query types. We hypothesise that this is because the ranking gives more control over the process, and items can be visually compared to previous items. We do not have a good intuition about why clustering was so unpopular. We expected this to be the easiest and most popular method, since jobs only had to be sorted into three categories (single best job/good jobs/bad jobs). 

\section{EXPERIMENTS III: APPLICATION} \label{sec:exp3-traffic}

In order to test our decision making tool in a real-world application, we worked on a project together with the municipality of Amsterdam. Part of their daily work is traffic control in the city, which is a complex real-world task with many objectives. For critical intersections in the city, the municipality utilises  specialised simulation software to test policies for traffic regulation. However, since there are many free parameters and each setting needs to be tested several times to account for variance in the simulation, it is near impossible for humans to manually test all settings. We thus hypothesise that policy makers can benefit from a decision-making support system like ours.

For our experiment, we gathered results from the simulation software (for $256$ different parameters settings, averaged over $10$ runs each), simulating two hours during rush hour for one of the city's busiest traffic areas. There are $11$ objectives, which together reflect the delay duration and queue length in different directions and for different traffic participants. First, we removed the Pareto-dominated results, leaving us with $75$ parameter settings with possibly optimal value.
Two experts from the municipality then used our web-interface as a tool to find the best outcome, according to their intrinsic utilities. In this case, we have no access to the true utility values of the users. Our experiments therefore investigate whether the experts think the system is useful when used in a real-world problem, and which preference elicitation strategy they prefer.

We only used pairwise and ranking queries in this experiment, since clustering performed poorly in our user study (sec \ref{sec:exp2-user}). Each participant started with one of the two query types, and then switched to the other. Instead of having a time limit, the participants now had the choice of either requesting a new item, or finishing the experiment by saying they are satisfied with the current outcome.

Table \ref{table:traffic-results} shows the results from this experiment. Both experts answered about twice as many queries for pairwise comparisons than for ranking before they were satisfied with the outcome, and both spent more time on the first experiment than on the second, regardless of query type. Both experts preferred the ranking query types. They both found the same item in the ranking query, but different ones in the pairwise comparison.
We received positive feedback from the experts, who found our multi-objective decision support system to be a useful tool. They preferred the ranking queries since it makes it easier to compare items.

We thus conclude that our multi-objective decision support system -- using ranking, a temporary prior heuristic, and virtual comparisons -- can be used by policy makers to aid them in their work.

\begin{table}
  \centering
  \begin{tabular}{|l|c|c|}
    \cline{2-3}
    \multicolumn{1}{c|}{}   & Expert 1 & Expert 2 \\ \hline
    Order 					& pairwise - ranking   & ranking - pairwise  \\ \hline
	\# Queries: pairwise 	& 22   	& 20  \\ 
    \# Queries: ranking		& 8 	& 12  \\ \hline
    Time (min): pairwise 	& 4:06   & 2:37  \\
    Time (min): ranking 	& 1:43   & 6:14  \\ \hline
    Preference				& ranking & ranking \\ \hline
\end{tabular}
\caption{Results in terms of how many items the experts looked at, how long each experiment took, and which query type they preferred.} \vspace{-0.8cm}
\label{table:traffic-results}
\end{table}

\section{RELATED WORK}

In this paper we model the utility function using GPs, which has important advantages: it explicitly models the uncertainty over the function, enabling active learning, and in contrast to e.g., using POMDPs or predecessing methods \cite{chajewska2000making, boutilier2002pomdp}, it can handle a continuum of items. 
The main framework for relative preferences in Gaussian processes we use is taken from the work of \citet{chu2005preference}.
They propose a likelihood for pairwise comparisons that can be used in GPs for preference learning. Their experiments are done on fixed datasets, i.e., they do off-line instead of active learning. The extension of these relative preference GPs to active learning by \citet{brochu2008active,brochu2010tutorial} asks the user to compare two items in each query; the winner from the last query and a new item suggested by the acquisition function. \citet{chu2005extensions} also propose an active learning approach with relative preference GPs, but for the learning to rank task. They, at each step, select the \emph{pair} of items with the highest expected entropy term. The objective in learning to rank is to approximate the unknown function, instead of finding the best item.
\citet{jensen2011efficient} extend the approach of \citet{chu2005preference} so that the user can in addition specify a degree of preference when comparing two items, for which they propose a novel likelihood to use with the Gaussian process. 
While this idea is in principle orthogonal to ours, due to the intrinsically ordinal nature of human preferences, we consider the risk of misspecification by introducing a ratio scale for degrees of preferences too high to attempt integrating it with our method. 
Pairwise preferences with GPs have been applied to a variety of settings, such as sound quality reduction mechanisms in hearing aids \cite{groot2011predicting}, material design \cite{brochu2008active,brochu2007preference} and learning to predict emotions expressed in music \cite{madsen2012predictive}. We believe that our strategies could be useful in many of these.

\section{CONCLUSION}

In this paper, we proposed new strategies to elicit user preferences in an active learning setting with Gaussian processes, by asking the user to rank or cluster items. We showed that compared to previous work which uses pairwise comparisons, ranking and clustering methods can both find a better approximation to the utility function, and select items with higher utility for the user in fewer queries. Furthermore, we showed that we can obtain better results by entering monotonicity information into the GP via virtual comparisons. We further showed that utilising a heuristic linear prior mean function during the first queries leads to increased performance, but that this heuristic should be turned off in later queries. 

A user study showed that the ranking method was most popular, although the perceived effort was slightly higher, and users felt less often understood by this method than for example the pairwise comparisons. Surprisingly, clustering methods were least popular. 

Finally, we successfully used our proposed decision-making tool in a real-world project on traffic regulation together with the municipality of Amsterdam. We showed how human expertise can successfully be combined with decision-making tools, and received positive feedback from our collaborators. We hope that this work inspires new real-world applications involving multiple objectives, and will be used in multi-objective optimisation \cite{coello2007evolutionary,deb2005scalable,halffmann2017general, liefooghe2017fitness}, planning \cite{benabbou2017adaptive,roijersPhD,wang2012multi,wray2015multi}, and RL \cite{auer2016pareto,mannion17,roijers2017adt,wiering2014model} alike.

In future work, we aim to perform more extensive real-user experiments.
We aim to extend our approach to multi-user settings, like has been done by \citet{houlsby2012collaborative} for pairwise feedback from multiple users or \citet{guo2010gaussian} who, for a new user, use an informed GP prior inferred from other. 
We also aim to study acquisition functions that pick more than one next item, based on look-ahead information gain \cite{boutilier2002pomdp}. Furthermore, we are interested in investigating a more principled approach to deciding when to disable the linear prior mean function.

\begin{acks}

The authors would like to thank the municipality of Amsterdam for their time and insights, and to the employees that tested our decision making tool. We thank the anonymous MTurk workers for participating in our study. This research was supported by Innoviris, the Brussels Institute for Research and Innovation, Brussels, Belgium. Diederik M.~Roijers is a postdoctoral fellow of the Research Foundation - Flanders (FWO), grant number 12J0617N.

\end{acks}

%%%%%%%%%%%%%%%%%%%%%%%%%%%%%%%%%%%%%%%%%%%%%%%%%%%%%%%%%%%%%%%%%%%%%%%%%%%%%%%%%%%%%%%%%%%%%%%%%%%%%%%%%
%% bibliography: see CFP for number of permitted pages

\bibliographystyle{ACM-Reference-Format}  % do not change this line!
\bibliography{pref-elicit}  % put name of your .bib file here

\end{document}